# Artificial Intelligence Approaches

Yingjie Hu[1], Wenwen Li[2], Dawn Wright[3], Orhun Aydin[3], Daniel Wilson[3], Omar Maher[3], Mansour Raad[3]

*[1]Department of Geography, University at Buffalo, Buffalo, NY 14260*
*[2]School of Geographical Sciences and Urban Planning, Arizona State University, Tempe, AZ 85287*
*[3]Esri Inc., Redlands, CA 92373*

**Summary Abstract:** Artificial Intelligence (AI) has received tremendous attention from academia, industry, and the general public in recent years. The integration of geography and AI, or GeoAI, provides novel approaches for addressing a variety of problems in the natural environment and our human society. This entry briefly reviews the recent development of AI with a focus on machine learning and deep learning approaches. We discuss the integration of AI with geography and particularly geographic information science, and present a number of GeoAI applications and possible future directions.

## Definitions

1. **Artificial Intelligence**: The study and design of machines or computational methods that can perform tasks that normally require human intelligence.
2. **GeoAI**: An interdisciplinary field of geography and artificial intelligence.
3. **Machine Learning**: A sub field in AI that relies on statistical methods or numerical optimization techniques to derive models from data without explicitly programming every model parameter or computing step.
4. **Deep Learning**: A special type of machine learning that leverages multiple layers of nonlinear processing units, or neurons, to learn representations from raw data to achieve the goal of automatic learning for completing various AI tasks.

## Description/body

## 1. AI and Geography

Artificial Intelligence (AI) has received tremendous attention in recent years from academia, industry, and the general public. Despite its recent popularity, the field was born back in 1956 at a workshop at Dartmouth College (McCarthy 1956). AI is a broad field from its beginning, and has many different definitions (Russell et al. 2003). Some definitions focus on designing intelligent machines that can act like humans. For example, the famous *Turing Test* was designed to see if the responses of a machine can be indistinguishable from those of a real person (Turing 1950). Some other definitions focus on designing and developing computational methods to complete tasks that typically require human intelligence, such as recognizing objects from images or understanding the meaning of natural language sentences. This entry is primarily based on the second type of definitions.

The development of AI has experienced falls and rises. Following its early optimism in 1960s and 70s, AI research went through the "AI winter" due to the failures of AI methods in addressing real-world problems. The following decades witnessed several other waves of optimism and disappointment. Since the 21th century, and especially after 2010, there has been significant progress in AI research. Three major factors have contributed to this fast advancement of AI: big data, novel algorithms, and immense computational power. The emergence of ubiquitous sensors and user-generated content on the Web allows large amounts of data to be generated and collected at a rapid pace. Big data enables computers to "observe" many different aspects of the world, to learn the ways in which the world functions, and to predict the future based on existing observations. Meanwhile, novel algorithms and models have been developed, and the AI community has embraced various ideas and theories from other fields, such as statistics, economics, biology, and cognitive science, in addition to its tradition of logics. Third, high performance computing (HPC) provides the essential power for linking big data and new computational models, and allows the training of sophisticated models on large datasets to be completed within hours or days rather than weeks or months. These three major factors, namely big data, novel algorithms, and immense computational power, greatly fueled the remarkable development of AI in recent years.

In this context, it is probably unsurprising to see the integration of AI and geography, particularly GIScience and GI Systems (GIS). A high volume of data nowadays contains georeferenced information, that is, information about locations on or near the surface of the Earth. Examples include GPS trajectories, remote sensing images, location-based social media, spatial footprints of buildings, roads, and parcels, global elevation data, land use and land cover data, population distribution, and so forth. These georeferenced data are critical inputs for many models that address a wide range of problems related to our human society and the natural environment, and GIS is essential for effectively processing, managing, and visualizing these big geo data. In addition, many AI models need to synthesize heterogeneous data from different sources, while geographic location is often the only factor that can link such heterogeneous datasets. Finally, existing research has already demonstrated successful integrations between GIS and HPC (Wang 2010), and integrated systems of GIS and AI also have the capability of leveraging high computational power.

Many efforts have already been devoted by both academia and industry to facilitating the integration of geography and AI, and the outcome is an exciting new and interdisciplinary area-- GeoAI. In academia, the first GeoAI workshop was held in November 2017 in a major GIS conference, ACM SIGSPATIAL, which attracted more than 100 participants (Mao et al. 2018a). In April 2018, the first AI and Deep Learning Symposium was held in the largest geography conference, Annual Meeting of American Association of Geographers (AAG). In industry, one of the major GIS companies, Esri Inc., has been collaborating with Microsoft to offer the GeoAI Data Science Virtual Machine (DSVM) with the goal of bringing AI, cloud computing, geospatial analytics, and geo-visualizations together. It is also worth noting that there are already studies reported in the literature that leverage machine learning for geographical problem-solving before the GeoAI workshop, and many of which, such as hyperspectral image analysis (Chen et al. 2014), high-resolution satellite image interpretation (Zhang et al. 2015), and 3D reconstruction (Blaha et al. 2016), are from the remote sensing community (Tuia et al. 2009, Zhu et al. 2017). Given this continuing interest, it is likely that there will be more interactions between geography and AI in the foreseeable future. This entry aims to provide interested readers, especially students who will become future GIS leaders, with some background of GeoAI. This field is rapidly developing, and this entry, completed between middle 2018 and early 2019, reflects the progress so far. New advancements are likely to happen in the following years.

## 2. Main AI Approaches

AI, as a broad field, encompasses many different approaches ranging from top-down knowledge representation to bottom-up machine learning. There are three related concepts that have been frequently used in recent years: *AI*, *machine learning*, and *deep learning*. In general, *AI* is the broadest concept, *machine learning* is a sub field in AI, and *deep learning* is a special type of machine learning. Figure 1 illustrates the relations among these three concepts. While the broad field of AI includes many approaches, its recent popularity is largely due to the outstanding performances of machine learning, especially deep

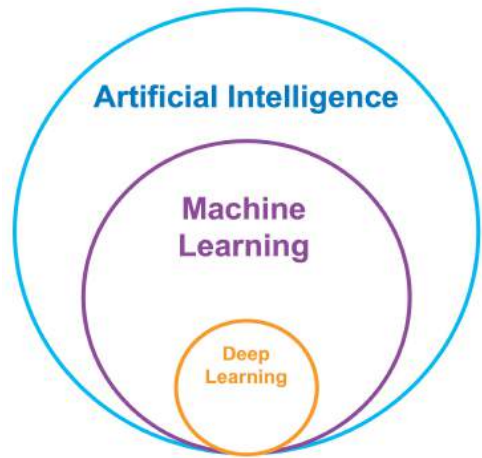

**Figure 1.** Relations among AI, machine learning, and deep learning (Bennett 2018).

learning. Therefore, this entry focuses on discussing these two types of AI approaches.

## 2.1 Machine learning

Machine learning is a sub field in AI (see Figure 1) that usually relies on statistical methods or numerical optimization techniques to derive models from data without explicitly programming every model parameter or computing step (Valiant 1984). One important characteristic shared by many machine learning models is the use of probability to represent the uncertainty that widely exists in real-world problems. There are three main types of learning: supervised learning, unsupervised learning, and reinforcement learning. Supervised learning requires labeled data for training a computational model, while unsupervised learning examines unlabeled data to discover patterns. Reinforcement learning does not require labeled data but needs action-based feedbacks, such as rewards or punishments, to help a computational model to learn. Machine learning tasks can be categorized in various ways. Based on their goals, we can identify tasks such as classification, clustering, and prediction (Bennett 2018). In classification, the goal is to classify a target into a category, e.g., classifying a land parcel into a category such as *Commercial* or *Agricultural*. In clustering, the goal is to detect clusters from data, e.g., finding the clusters of vehicles based on their locations to detect traffic congestions. In prediction, the goal is to predict unknown values, e.g., predicting the average temperatures of several locations in the near future based on their historical temperatures and other variables using a regression model. There are also other tasks that can fall under the umbrella of machine learning, such as anomaly/novelty detection, data generation, visualization, feature learning, and others. A variety of machine learning models have been developed, such as regression, decision tree, random forest, support vector machine (SVM), naïve Bayesian classifier, density-based clustering, hidden Markov model (HMM), artificial neural network (ANN), and numerous others. These methods are discussed with details in machine learning textbooks, such as Flach (2012). While most machine learning methods can be directly applied to geographic data, they typically do not take into account the uniqueness of geographic phenomena, such as spatial autocorrelation and spatial nonstationarity. There exist some methods, such as Empirical Bayesian Kriging (EBK) regression (Krivoruchko 2012) and spatial Principal Component Analysis (sPCA) (Jombart et al. 2008), that explicitly model the spatial aspect of geographic problems by, e.g., including spatial weights. Some classic spatial models, such as geographically weighted regression (GWR) (Fotheringham et al. 2003), can also be used in a machine learning manner, i.e., they can be firstly trained on one dataset and then tested on other datasets.

## 2.2 Deep learning

Deep learning is a special type of machine learning that focuses on developing and using deep neural networks (DNN) for machine learning tasks. DNN is a special type of artificial neural network which has multiple layers (also called hidden layers) between the input and the output layers. Each layer consists of a collection of computing units, called neurons, which take the input from the previous layer and generate a non-linear output to the next layer. Deep learning has gained a huge amount of interest in recent years due to its outstanding performances (LeCun et al. 2015), thanks to the availability of large labeled datasets, such as ImageNet (Deng et al. 2009), and HPC. Similar to other machine learning models, deep learning can be utilized to complete tasks in classification, clustering, prediction, and so forth. Particularly, two types of DNN, convolutional neural network (CNN) and recurrent neural network (RNN), have received a lot of attention from the geography community. CNN is especially suitable for processing images by extracting and representing abstract features through a cascade of neuron layers and using convolutional filters (Li et al. 2017a, Maggiori et al. 2017, Gong et al. 2018). RNN is suitable for processing sequence data, such as movement trajectories (which can be modeled as a sequence of locations) (Kulkarni and Garbinato 2017), by memorizing some of the previous states and establishing links between the current and previous states. While many studies applied existing models to geographical problems, researchers also developed new DNN models specifically for handling geographic data. Marcos et al. (2018) proposed Rotation Equivariant Vector Field Network (RotEqNet) for land cover mapping based on remote sensing images. RotEqNet encodes rotation equivariance in a CNN, and can recognize the rotated versions of the same object from remote sensing images while reducing the number of parameters required by a traditional CNN. Srivastava et al. (2018) proposed a Variable Input Siamese Convolutional Neural Network (VIS-CNN) model for classifying urban-object level land use types. Their VIS-CNN model can accept a variable number of Google Street View images for an urban object and aggregate them to learn the land use type in an end-to-end manner.

## 3. Applications of AI in Geography

There exist a considerable number of applications of AI in the domain of geography. This section summarizes some of these applications.

### 3.1 Automatic recognition of natural terrain features from remote sensing imagery

Natural terrain features, such as craters, volcanos, and sand dunes, are important indicators of the Earth geological process. Detecting where they are and extracting their geomorphological properties are of great importance to geographers and geologists in understanding the formation process of different terrain features, differentiating similar landscapes, as well as enriching our geospatial knowledge. While this task has been done previously primarily using Object-Based Image Analysis (OBIA), it can hardly achieve automation of the processing steps. Parameters, such as scale factor, and the merging of segmented super pixels (a cluster of pixels with similar values) are often done manually or semi-automatically. With the proliferation of spatial big data, such as massive remote sensing imagery, as well as the fast evolution of deep learning techniques, it becomes possible for implementing an automatic learning process to detect and characterize different terrain features.

Li et al. (2017b) extended a Faster-RCNN (Faster Region-based CNN) model to support the automatic detection of natural terrain features from remote sensing imagery. Different from the classification tasks using CNN, object detection not only needs to tell what is inside of an image, but also the location of the object, depicted by a bounding box (BBOX). This requires additional location labels in the training data as well. Several challenges were addressed in applying deep learning to such spatial problems. This includes a lack of training database, the ambiguous boundary of natural features in comparison to man-made features, such as building or golf courses, and the inter-category similarity of different features. For instance, there is usually a crater type of feature on top of a volcano, called "volcanic crater", which looks similar to impact craters. Making such distinctions may require the deep learning network to learn features that is unique to each (sub)category, such as emitting ashes from active volcano and flatter terrain that an impact crater usually appears in. Based on a test database of over 10,000 images, the deep learning network achieves over 90% of mean average precision (mAP) for the task of detecting eight terrain categories (Li and Hsu 2018). The derived information, both category and BBOX, can complement the limited spatial information which is only a point for representing features of any extent in known gazetteers, such as the USGS (United States Geological Survey) GNIS (Geographic Name Information System), hence providing better support for feature allocation and landscape interpretation. Figure 2 demonstrates detection results for hill, impact crater, meander, and volcano.

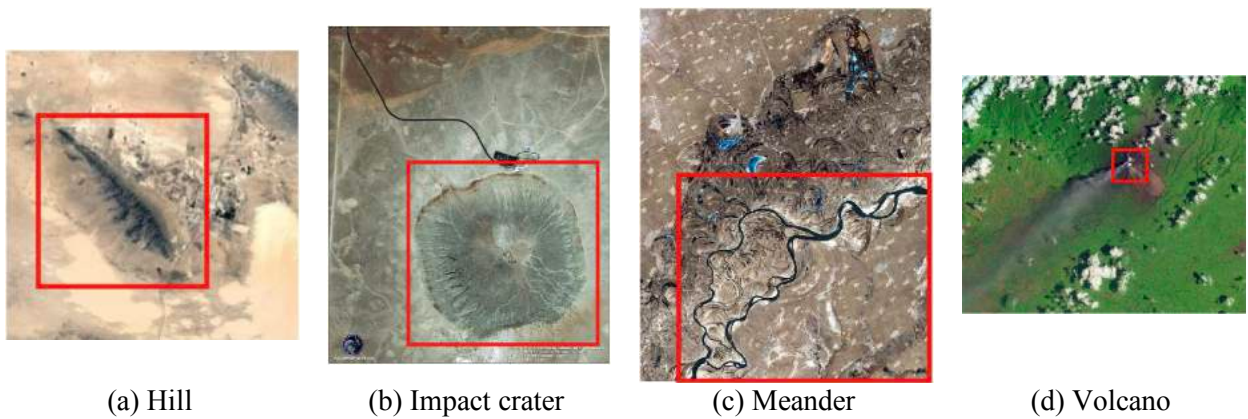

(a) Hill        (b) Impact crater        (c) Meander        (d) Volcano

**Figure 2**. An illustration of terrain feature detection results of hill (a), impact crater (b), meander (c), and volcano (d) from remote sensing imagery.

*3.2 Land cover classification for conservation*

Another excellent example of AI application in geography is the work of the Chesapeake Conservancy, a non-profit organization based in Annapolis, MD, and a pioneer in the field of precision conservation to monitor, protect, and restore the natural environment. Their "mission field" covers the entire Chesapeake Bay watershed, supports more than 3,600 species of plants and animals, and is home to 17 million people. Despite the importance of the Chesapeake Bay watershed, its health over decades has been marred by sewage overflows and runoff of animal waste and chemicals. The future of the watershed depends upon smart conservation that is informed by data about where growth is least harmful and where interventions can be most helpful. This requires detailed and highly accurate land cover maps to understand how land is used and to identify environmental issues, such as pollutants entering the bay through agricultural runoff. While Landsat imagery offer a standard 30-meter resolution, they are not sufficient for achieving full understanding of all the possible scenarios of 21st-century land use and associated disruptions of nature. Land cover data at 1-m resolution are not yet available everywhere across the US and is often difficult and expensive to obtain, much less to process and interpret, especially for less affluent communities, smaller nonprofits, and the like. A driving motivation of the Chesapeake Conservancy is to empower natural resource managers and conservationists with access to such a high-resolution data resource.

To achieve this goal, the Chesapeake Conservancy took a deep learning approach to create a semantic segmentation (pixel-level classification) model which predicts high-resolution land cover from aerial imagery (Allenby et al. 2018). The imagery was obtained from the National Agriculture Imagery Program (NAIP), at a 1-m or 60-cm resolution. Figure 3 illustrates this approach based on a location within the Chesapeake watershed. In the upper left is NAIP imagery from the Esri Living Atlas of the World (Kensok 2017); in the upper right is the original Chesapeake land cover map; and then immediately below that in the lower right is the map as produced by a deep residual neural network executed on that NAIP imagery, on the fly, classifying forests, fields, water, and impervious surfaces such as roads and houses; in the lower left is the same region, except it's a mixture of probabilities from the model, across all those categories, currently at 91% accuracy as compared to ground observations. This was extremely useful as it communicates the uncertainty in the detections and allows one to see how the algorithm was "thinking" and what was needed to improve it. The real power of this approach is that data scientists can use this same algorithm to classify land cover in places that it has never seen before. The system can classify at a rate significantly faster than manual methods, limited only by the available hardware. The time needed to correctly classify the entire Chesapeake Bay watershed was reduced from 2500 hours to 150. The goal is to take this AI approach to scale across all the other watersheds of the US, and to empower more organizations in more places to sustainably manage their lands.

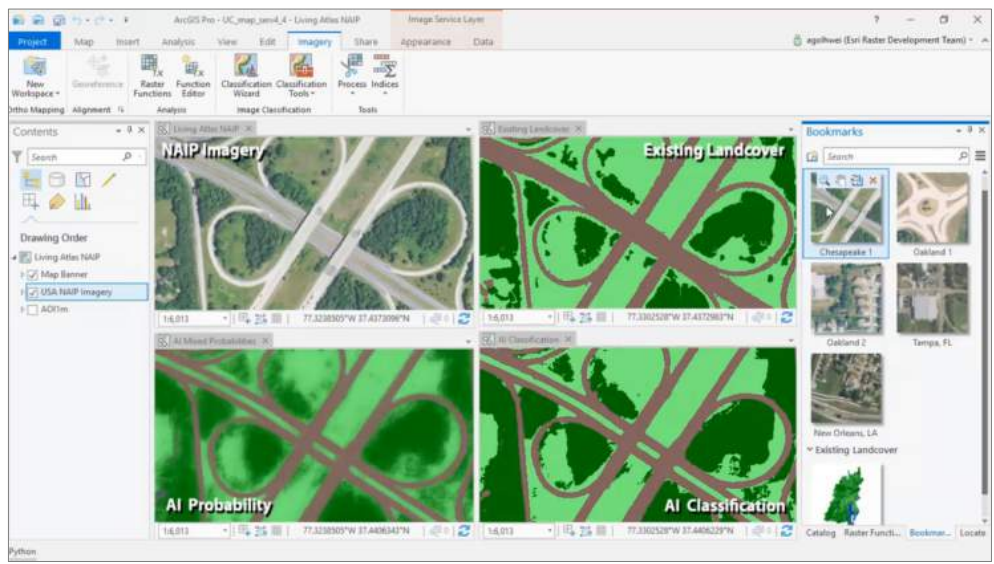

**Figure 3.** High-resolution land cover classification of Chesapeake Bay watershed using a deep residual neural network implemented in ArcGIS Pro.

### 3.3 Modelling seagrass habitats in space and time

In this AI application, GIS and machine learning methods are integrated to model the relation between seagrass habitats and ocean conditions. Seagrasses are marine plants that can quickly sequester vast amounts of $CO_2$, up to 100 times more and 12 times faster compared to tropical forests (Parry et al. 2007, Pidgeon 2009). Only limited amounts of data on global seagrass habitats are available, and existing data are often spatially sparse. Therefore, developing a seagrass habitat model based on existing data can help quantitatively understand the ocean conditions that favor the growth of seagrass. In addition, such a model can also predict future seagrass habitats based on changing ocean conditions. Multiple data sources, including the seagrass data from MarineCadastre.gov and the recently available Ecological Marine Units (EMU) dataset (Wright et al. 2017) were used in this analysis. These data were employed to train a model based on the seagrass occurrence along the U.S. coast to predict seagrass habitats globally up to a depth of 90 meters. In situ oceans data are interpolated using Empirical Bayesian Kriging (EBK) to produce global-scale predictions. A random forest framework is then used to relate the occurrence and absence of seagrass to a set of variables including temperature, salinity, phosphate, silicate, nitrate, and dissolved oxygen. Figure 4(a) shows the global-scale predictions on seagrass habitats using U.S. coast data for training, which is juxtaposed against a map in Figure 4(b) by Short et al. (2007) of seagrass observations.

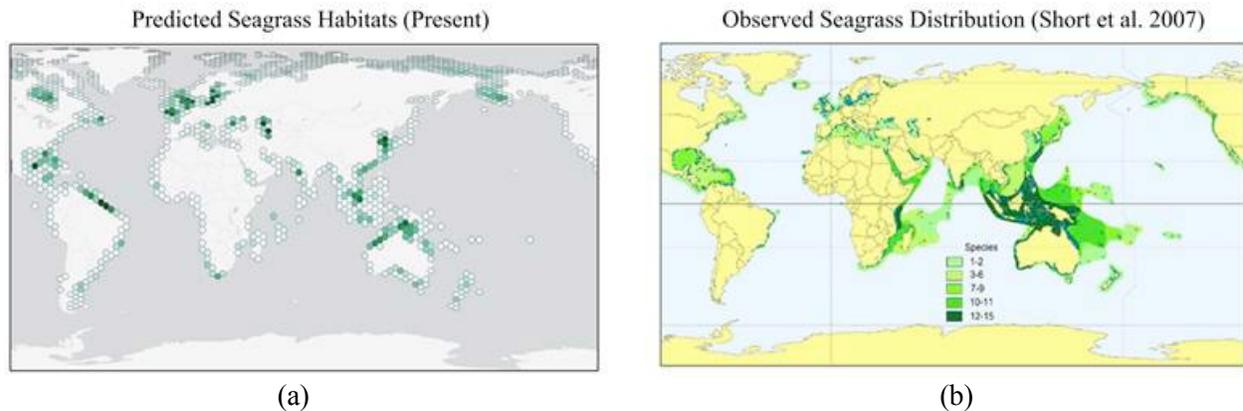

(a)                                                                    (b)

**Figure 4**. (a) Predicted seagrass habitats using a random forest model (dark green indicates high modelled abundance); (b) reported seagrass occurrence data by Short et al. (2007).

The random forest model trained on existing seagrass habitats is then applied to different ocean conditions to forecast the possible scenarios of seagrass habitats due to the warming of the oceans. An increase of 2 ℃ in average ocean temperature is simulated with an increment of 0.2 ℃. For every snapshot of ocean condition, seagrass habitats are predicted using the trained random forest model. Getis-Ord Gi* statistic (Ord and Getis 1995) is used for detecting the clusters of seagrass abundance over space-time bins. Assessing changes in the intensity of abundance at individual locations over time using the Mann-Kendall statistic yields the emerging hot spot map in Figure 5.

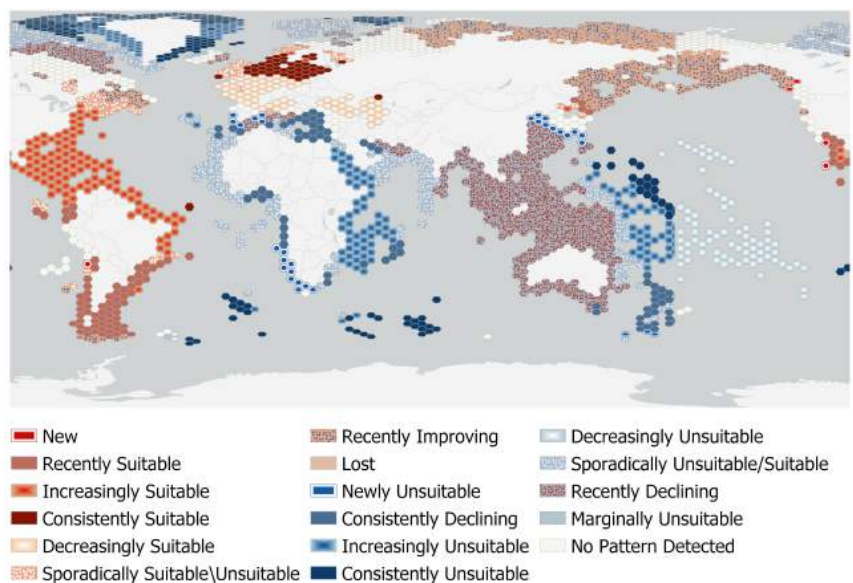

**Figure 5**. *Emerging hot spot map for seagrass habitats under increasing ocean temperature.*

As shown in the figure, Australia could lose its seagrasses under changing ocean conditions, whereas Siberia coast may improve its suitability for seagrass habitats. The use of random forest in this AI application enables a data-driven model of seagrass habitats for world's coasts. A tutorial demonstrating the prediction of seagrass with machine learning methods can be found in Additional Resource 5.

*3.4 Summary and some other applications of AI in geography*

The three studies discussed above are summarized in Table 1.

Table 1. A summary of the three studies discussed from 3.1 to 3.3.

| *Study* | *Task Type* | *Learning Type* | *Model(s)* |
|---|---|---|---|
| Automatic recognition of natural terrain features | Classification | Supervised learning | Deep learning: faster region-based CNN |
| High-resolution land cover classification | Classification | Supervised learning | Deep learning: deep residual neural network |
| Global-scale seagrass habitat prediction | Prediction | Supervised learning | Machine learning: EBK and random forest |

As shown in Table 1, the first study addresses a classification task (more specifically, object detection) which leverages deep learning to identify the category of a terrain feature and its location in the image. The second study is also a classification task (semantic segmentation), in which a pixel-level land cover classification is achieved. The third study completes a prediction task which predicts the growth of seagrass in simulated situations. All three studies employ supervised learning in which the models are first trained using labeled training data. While supervised learning is indeed very common, many studies also employ unsupervised learning, such as spatial and spatio-temporal clustering approaches, to extract geographic patterns from data (Anbaroglu et al. 2014, Hu et al. 2015).

There exist other applications that integrate AI techniques with geographic research. CNNs and their variants, in addition to their outstanding performances in processing satellite imagery (Collins et al. 2017) and Google Street View photos (Law et al. 2017), are also utilized for aligning vector data to raster historical map images (Duan et al. 2017), detecting the types of online maps (Zhou et al. 2018), classifying geographic texts (Adams and McKenzie 2018), and so forth. RNNs, such as Long Short-Term Memory (LSTM), are utilized to handle time series data, such as predicting the next locations of trajectories (Li et al. 2018) and examining the temporal patterns of crops (Sun et al. 2018). RNNs are also used for analyzing geotagged tweets and other natural language texts containing geographic information (Mao et al. 2018b, Sit et al. 2019, Santos et al. 2018). Machine learning models, such as hidden Markov model, are integrated with a variety of geospatial applications, such as indoor navigation (Li et al. 2017a) and location prediction of financial services (McKenzie and Slind 2019). While many applications already exist, new ideas and methods are being constantly developed.

## 4. Possible Future Directions

GeoAI is a rapidly growing field with many possible directions to be pursued in the near future. Here, we list a few of these directions. First, most AI methods are currently applied to pre-defined and highly specific spatial analysis tasks. Can there be a general GeoAI assistant, similar to Amazon's Alexa or Apple's Siri? Such a GeoAI assistant may be able to understand the needs of a GIS practitioner, automatically formalize and define the tasks, and identify candidate tools from a large GIS toolbox. Second, most of today's AI models are developed based on a training dataset, and as a result, they naturally inherit the potential bias in the data. In geographic research, training data are often collected from a certain geographic area, and consequently, it can be difficult for a model trained using the data from one geographic area to perform well on the data from other areas. An important direction, therefore, is to improve model architectures (or the training process) so that the obtained GeoAI models can be transferred across different geographic areas. Third, many existing GeoAI studies only apply AI methods to geographic problems rather than improving or inventing methods. While this is fine from a perspective of problem solving, it is critical for geographers to not only *import* methods from outside disciplines but also *export* geographic knowledge to other fields. Geographically-informed or spatially-explicit AI models can be developed to capture the uniqueness of geographic problems.

## Additional Resources

1. GeoAI Data Science Virtual Machine – http://esriurl.com/geoai2018
2. Microsoft AI for Earth Initiative including grants – http://aka.ms/aiforearth
3. AI for Earth Deep Learning Student Story Map – http://esriurl.com/cassava
4. Machine Learning Tools in ArcGIS – http://esriurl.com/ml
5. Learn ArcGIS Lesson – Predict Seagrass with Machine Learning - https://learn.arcgis.com/en/projects/predict-seagrass-habitats-with-machine-learning/
6. ArcGIS Export Training Data for Deep Learning Tool – http://esriurl.com/dltool
7. Podcast – Location Intelligence + Artificial Intelligence: Making Data Smarter, Part 1 - https://www.esri.com/about/newsroom/podcast/location-intelligence-artificial-intelligence-making-data-smarter/
8. Podcast – Location Intelligence + Artificial Intelligence: Making Data Smarter, Part 2 - https://www.esri.com/about/newsroom/podcast/location-intelligence-artificial-intelligence-making-data-smarter-part-2/
9. Podcast – How AI and Location Intelligence Can Drive Business Growth - https://www.esri.com/about/newsroom/podcast/ai-and-location-will-drive-tomorrows-digital-transformations/
   *All resources above are accessible on Feb. 28th, 2019

## Learning Objectives

1. Explain the general concept of artificial intelligence.
2. Describe the relations among AI, machine learning, and deep learning.
3. Explain the main difference between supervised learning and unsupervised learning.
4. Describe an example application of AI in geography.

## Instructional Questions

1. What is AI and how is AI related to machine learning and deep learning?
2. Why geography and GIS are critical for AI to address many of the real-world problems?
3. What are some machine learning models that capture the uniqueness of geographic phenomena?
4. What is the main difference between supervised learning and unsupervised learning?